\documentclass[10pt,a4paper,twocolumn]{article}
\usepackage{graphicx} % Required for inserting images
\usepackage{times}
\usepackage[a4paper, total={6.5in, 10in}]{geometry}
\usepackage{blindtext}
\usepackage{amssymb}
\usepackage{tabto}
\usepackage{amsmath}
\usepackage{physics}
\usepackage{mathtools}
\usepackage{caption}
\usepackage{supertabular}
\usepackage{fancyhdr}
\usepackage{xcolor}

\usepackage{titling}

\title{\textbf{ Expressive Symbolic Regression for Interpretable\\
 Models of Discrete-Time Dynamical Systems}}
\author{Adarsh Iyer$^{1\dag}$, Nibodh Boddupalli$^2$, Jeff Moehlis$^2$\\
        \fontsize{11}{12}\selectfont
        $^1$\textit{Lynbrook High School, 1280 Johnson Ave. San Jose, CA 95129}\\
        \fontsize{11}{12}\selectfont
        $^2$\textit{Department of Mechanical Engineering, University of California, Santa Barbara, CA 93106}\\
        \fontsize{11}{12}\selectfont
        $^\dag$\textit{Corresponding author: adarshramiyer@gmail.com}}
\date{}
\begin{document}

\setlength{\headheight}{40pt}

\fancypagestyle{firststyle}
{
   \fancyfoot{}
\pagestyle{fancy}
\fancyhead[L]{\fontsize{9}{10}\selectfont \color{gray} ITERATED MAP IDENTIFICATION}

\fancyhead[C]{\fontsize{10}{11}\selectfont UCSB RMP\\}

\fancyhead[R]{\fontsize{9}{10}\selectfont \color{gray} AUG 2023}
}

\fancyfoot{}
\pagestyle{fancy}
\fancyhead[L]{\fontsize{9}{10}\selectfont \color{gray} ITERATED MAP IDENTIFICATION}

\fancyhead[C]{\fontsize{10}{11}\selectfont UCSB RMP\\}

\fancyhead[R]{\fontsize{9}{10}\selectfont \color{gray} AUG 2023}

\twocolumn[{\centering{\maketitle}}
\thispagestyle{firststyle}
Interpretable mathematical expressions defining discrete-time dynamical systems (iterated maps) can model many phenomena of scientific interest, enabling a deeper understanding of system behaviors. Since formulating governing expressions from first principles can be difficult, it is of particular interest to identify expressions for iterated maps given only their data streams. In this work, we consider a modified \underline{Sym}bolic \underline{A}rtificial \underline{N}eural \underline{N}etwork-\underline{T}rained \underline{Ex}pressions (SymANNTEx) architecture for this task, an architecture more expressive than others in the literature. We make a modification to the model pipeline to optimize the regression, then characterize the behavior of the adjusted model in identifying several classical chaotic maps. With the goal of parsimony, sparsity-inducing weight regularization and information theory-informed simplification are implemented. We show that our modified SymANNTEx model properly identifies single-state maps and achieves moderate success in approximating a dual-state attractor. These performances offer significant promise for data-driven scientific discovery and interpretation.
\medbreak
\par\vspace{6ex}]

\section*{\centering{\fontsize{11}{12}\selectfont
I. INTRODUCTION}}
\newcommand{\x}{\mathbf{x}}
\newcommand{\f}{\mathbf{f}}
\newcommand{\R}{\mathbb{R}}

Discrete-time dynamical systems (iterated maps) of scientific interest, including comet trajectories [1, 2], particle dynamics in accelerators [3], optical perception of flickers [4], cardiac arrhythmias [5], and chaotic unforced oscillators [6], can be modeled by governing expressions of the form $\x_{t+1}=\f(\x_t)$, where $\x \in \R^n$ is a vector of $n$ state variables evolving over time $t$ as $\f : \R^n \rightarrow \R^n$. In many cases, limited knowledge of domain-specific mechanisms makes first-principles formulation of governing expressions infeasible. Moreover, chaotic behavior (\textit{e.g.} sensitivity to initial conditions and period-doubling bifurcations) arising from certain iterated maps can make traditional analyses intractable. Thus, modeling the dynamics of chaotic scientific systems based solely on their state-data streams is of particular interest. Formally, we seek a good estimate of $\f (\x) \equiv \x_2, \x_3, … \x_{t+1}$ given state data $\x \equiv \x_1, \x_2, … \x_t$.

Conventional machine learning methods can estimate such a function successfully but suffer from diminished insight into their internal processes. In contrast, we seek parsimonious expressions for $\f$ to aid interpretation. Such automated formulation offers immense promise for scientific discovery across disciplines. Naturally, some antecedent models for this task exist in the literature.

Brunton, Proctor, and Kutz [8] proposed the \underline{S}parse \underline{I}dentification of \underline{N}onlinear \underline{Dy}namics (SINDy) model. SINDy used sparsity-inducing linear regression on a library of candidate functions, such as trigonometric and polynomial transformations of state variables. Although this method achieved favorable results, aspects of the nonlinearities, \textit{e.g.} trigonometric periods and polynomial orders, remained fixed throughout training, leading to less expressive equations. In order to achieve more precise results without \textit{a priori} knowledge, a large number of candidate functions must be added to the library, a problem that becomes combinatorially intractable especially for systems of multiple state variables.

A more expressive approach to symbolic regression is that of Fronk and Petzold [9], who applied $\pi$-net V1 [10] within the Neural Ordinary Differential Equations (Neural ODEs) framework [11]. Their Polynomial Neural ODEs model was novel in its independence from SINDy [8] and was capable of learning and formulating polynomial expressions using trainable network weights. However, this model cannot capture other nonlinear functions (except through Taylor series approximations) and is restricted to integer polynomial orders. Moreover, as a consequence of using Hadamard product sublayers, an undesirable nesting phenomenon is apparent in the output equations.

Martius and Lampert introduced the \underline{Eq}uation \underline{L}earner (EQL) [12], a more general-purpose model. The EQL model augmented dense neural networks by incorporating nonlinearities (\textit{e.g.} $\sin(\cdot)$ and $\cos(\cdot)$) analogous to activation functions and two-element product units. Although this model successfully approximated physical systems, it limits polynomial orders to integers, with a maximum order dependent on the network depth. Thus, non-integer-order polynomials cannot be represented by EQL outputs. Moreover, the authors state that they cannot support division due to singularities.

In this work, we evaluate and examine a modified \underline{Sym}bolic \underline{A}rtificial \underline{N}eural \underline{N}etwork-\underline{T}rained \underline{Ex}pressions (SymANNTEx) model [7] for this task, particularly on the chaotic regimes of iterated maps. Nonlinear operators and weight-defined signomials make SymANNTEx more expressive than the previously detailed symbolic regression models and our modification creates more optimal fits. In the next section, we provide key details on this architecture, training techniques, and our modifications.

\section*{\centering{\fontsize{11}{12}\selectfont
II. METHODS}}
\subsection*{\fontsize{10}{11}\selectfont
\normalfont \textit{A. SymANNTEx Architecture}}
SymANNTEx [7] composes functions through $K$ stacks of $L$ operational layers. The input  is the state vector $\x$ at $M$ data points with a bias term appended, $(x, 2)^T \in \R^{n+1}$, and the output is $\f (\x) \in \R^n$. More details on the architecture, especially its expressiveness, can be found in the original paper [7]. The model is trained and simplified as outlined in the next subsection.

\subsection*{\fontsize{10}{11}\selectfont
\normalfont \textit{B. Initialization, Regularization, and Training}}
We use normal weight initialization with a standard deviation of $5\times 10^{-4}$ to avoid numerical problems. We follow the same $L_{1/2}$ [13], $L_{ops}$, and $L_{poly}$ regularization schemes as in [7]. The loss $\mathcal{L}$ is thus given by
\begin{equation}
    \mathcal{L} = \text{MAE} + \sum_{k=1}^{K} \sum_{l=1}^{L}(\alpha_1 L_{1/2}^{(k, l)} + \alpha_2 L_{poly}^{(k,l)} + \alpha_3 L_{ops}^{(k,l)})\text{,}
\end{equation}

where the regularization weights are $\alpha_1 = 0.05$, $\alpha_2 = 0.01$, and $\alpha_3 = 0.0375$, and $\text{MAE}$ is the Mean Absolute Error 
\begin{equation}
    \text{MAE} = \frac{1}{M}\sum_{m=1}^{M} \lVert \f (\x_{[m]}) - \Tilde{\f}(\x_{[m]}) \rVert _1
\end{equation}

where $\x_{[m]}$ is the $m^{\text{th}}$ data point. Motivation for these loss and regularization choices can be found in [7]. SymANNTEx is implemented in \textit{Pytorch} and optimized using Adam [14] with cyclic learning rate scheduling [15]. To promote generalizability, the network is trained for 5000 epochs in a five-fold cross-validation routine.

\subsection*{\fontsize{10}{11}\selectfont
\normalfont \textit{C. Simplification and Refinement}}
For interpretability, we assemble an expression representing the network through one forward propagation using \textit{sympy} [16]. With the goal of parsimony, we simplify the network equation at 11 logarithmically spaced rounding thresholds between $10^{-2}$ and $10^0$ by employing \textit{sympy}’s \textit{nsimplify} method. We select the threshold creating an expression $\Tilde{\f}_A$ minimizing the Akaike Information Criterion (AIC) [17], a classical measure of model quality.

We note that $\Tilde{\f}_A$ may not necessarily be optimally fit due to the removal of contributing terms. Assuming that the AIC simplification properly captures the form of the expression, we introduce a novel downstream Ordinary Least Squares (OLS) refinement phase in which we optimize the coefficients of $\Tilde{\f}_A$ to the data. In the next section, we outline our experimental procedures for a SymANNTEx model with downstream refinement.

\section*{\centering{\fontsize{11}{12}\selectfont
III. EXPERIMENTAL EVALUATION}}
\renewcommand{\figurename}{\fontsize{9}{10}\selectfont FIG. }
\renewcommand{\tablename}{\fontsize{9}{10}\selectfont TABLE }
\renewcommand\thetable{\Roman{table}}

We evaluate our modified SymANNTEx model on the chaotic regimes of the Logistic Map, Tinkerbell Map, and Gaussian map. To emulate natural imperfections, the data is perturbed with independently-drawn Gaussian noise as
\begin{equation}
    \x \vcentcolon = \x + \mathcal{N} \left( 0, \sigma \cdot \sqrt{\frac{1}{M} \sum_{m=1}^M (\x_{[m]})^2} \right) \text{,}
\end{equation}
where $\sigma$ characterizes the standard deviation of the noise relative to the Root Mean Square ($\text{RMS}$) of each state.

For each $\sigma$, we perform 20 instances of five-fold training and choose the best model in terms of validation-set $\text{MAE}$. The best model’s simplified and refined equation is evaluated through the Relative Root Mean Square Error 
\begin{equation}
    \text{RRMSE} = \sqrt{\frac{\sum_{m=1}^M \lVert \Tilde{\f}_A(\x_{[m]}) - \f(\x_{[m]})\rVert_2^2}{\sum_{m=1}^M \lVert \f (\x_{[m]}) \rVert_2^2}} \text{.}
\end{equation}

We note that even the ground truth expression produces non-zero RRMSEs for $\sigma > 0$, and models are thus compared to this performance rather than zero.
The best model out of all of the instances is chosen based on the validation-set RRMSE. Further experimental details are provided in Table I and the following three subsections.

\subsection*{\fontsize{10}{11}\selectfont
\normalfont \textit{A. Logistic Map}}

The logistic map finds relevance in scientific phenomena displaying period-doubling bifurcations [4, 5]. Therefore, we attempt to identify its governing expression
\begin{equation}
    x_{t+1}=rx_t(1-x_t) \equiv rx_t-rx_t^2
\end{equation}

for $r = 3.9$. We train SymANNTEx on a single trajectory initialized at $0.5$.

\subsection*{\fontsize{10}{11}\selectfont
\normalfont \textit{B. Gaussian Map}}

As an example utilizing the common operators, we experiment with the Gaussian map given by 
\begin{equation}
    x_{t+1}=e^{-\alpha x_t^2}+\beta
\end{equation}

for $\alpha = 12$ and $\beta = -0.5$. We train SymANNTEx on a single trajectory initialized at zero.

\subsection*{\fontsize{10}{11}\selectfont
\normalfont \textit{C. Tinkerbell Map}}

Iteration of the Tinkerbell Map gives rise to a strange attractor with chaotic dynamics [18]. As a more challenging example, we attempt to identify its governing expressions
\begin{subequations} 
\begin{align} 
x_{t+1}=x_t^2-y_t^2+ax_t+by_t\\
y_{t+1}=2x_t y_t + cx_t+dy_t
\end{align} 
\end{subequations}

for $a = 0.9$, $b = -0.6013$, $c = 2$, and $d = 0.5$. We train SymANNTEx on one trajectory initialized at $(-0.5, -0.5)$.

{\setlength{\tabcolsep}{1ex}
\begin{table}[h]
    \centering
    \begin{tabular}{cccccccc}
        \hline
        
        \fontsize{8}{9}\selectfont\textit{Map} & \fontsize{8}{9}\selectfont\textit{L} & \fontsize{8}{9}\selectfont\textit{K} & \fontsize{8}{9}\selectfont\textit{M} & \fontsize{8}{9}\selectfont\textit{Operators} & \fontsize{8}{9}\selectfont\textit{Params} & \fontsize{8}{9}\selectfont\textit{min. lr} & \fontsize{8}{9}\selectfont\textit{max. lr}\\
        
        \hline
        
         \fontsize{8}{9}\selectfont\textit{Logistic} & \fontsize{8}{9}\selectfont\textit{1} & \fontsize{8}{9}\selectfont\textit{1} & \fontsize{8}{9}\selectfont\textit{1000} & \fontsize{8}{9}\selectfont\textit{[sin, abs]} & \fontsize{8}{9}\selectfont\textit{16} & \fontsize{8}{9}\selectfont\textit{28e-3} & \fontsize{8}{9}\selectfont\textit{36e-3}\\

         \fontsize{8}{9}\selectfont\textit{Gaussian} & \fontsize{8}{9}\selectfont\textit{1} & \fontsize{8}{9}\selectfont\textit{2} & \fontsize{8}{9}\selectfont\textit{1000} & \fontsize{8}{9}\selectfont\textit{[exp]} & \fontsize{8}{9}\selectfont\textit{44} & \fontsize{8}{9}\selectfont\textit{36e-3} & \fontsize{8}{9}\selectfont\textit{48e-3}\\

         \fontsize{8}{9}\selectfont\textit{Gaussian} & \fontsize{8}{9}\selectfont\textit{2} & \fontsize{8}{9}\selectfont\textit{2} & \fontsize{8}{9}\selectfont\textit{1000} & \fontsize{8}{9}\selectfont\textit{[sign, sin]} & \fontsize{8}{9}\selectfont\textit{158} & \fontsize{8}{9}\selectfont\textit{36e-3} & \fontsize{8}{9}\selectfont\textit{48e-3}\\

         \hline
         
    \end{tabular}
    \caption{\fontsize{9}{10}\selectfont Hyperparameters for the Logistic, Gaussian, and Tinkerbell Map experiments. Params is the count of trainable network weights for each model and min. lr and max. lr are the constants used for cyclic learning rate scheduling with Adam.}
    \label{tab:my_label}
\end{table}}

\section*{\centering{\fontsize{11}{12}\selectfont
IV. RESULTS}}
We presented the scientifically pertinent task of dynamical system identification in Section I and outlined our modified SymANNTEx [7] pipeline for this task in Section II. In this section, we present our identified models and expressions for the maps in Section III. We characterize SymANNTEx for each of these maps and discuss useful techniques for its pragmatic employment.

\subsection*{\fontsize{10}{11}\selectfont
\normalfont \textit{A. Logistic Map}}
Our modified SymANNTEx model identifies
\begin{subequations} 
\begin{align} 
x_{t+1}=3.7392x_t-3.7578|x_t|^{2.044}+0.016 \text{,}\\
x_{t+1}=3.874x_t-3.8735|x_t|^{2.0094} \text{,}\\
x_{t+1}=4.1303_t-4.1021|x_t|^{1.8832} \text{,}
\end{align}
\end{subequations}

at $\sigma = 0.0$, $0.01$, and $0.05$, respectively. Further details are given in Table II. These expressions, particularly Eq. (8b), display a remarkable similarity to the true model. Moreover, Fig. 1b shows that even under chaotic sensitive dependence, iterations of Eqs. (8a) and (8b) closely follow the true model for 19 and 14 timesteps respectively. 

Fig. 1c demonstrates the reduction in RRMSE due to our novel OLS refinement stage. As contrasted to the $\sigma = 0.0$ and $\sigma = 0.01$ cases where the achieved RRMSE is slightly higher than the true RRMSE, Eq. (8c) and some other expressions with $\sigma = 0.05$ achieve RRMSEs better than that of the true model, a discrepancy also shown in Table II. In this case, the true function is not optimal due to the high $\sigma$, so Eq. (8c) fails to capture the correct expression and rapidly diverges from the true model’s trajectory (see Fig. 1b). Therefore, extra preprocessing steps should be taken to ensure better model fitting when $\sigma \geq 0.05$.

\subsection*{\fontsize{10}{11}\selectfont
\normalfont \textit{B. Gaussian Map}}
Our modified SymANNTEx model identifies
\begin{subequations} 
\begin{align} 
x_{t+1}=1.1e^{\left( -7.444|x_t|^{1.7}\right)}-0.571 \text{,}\\
x_{t+1}=-2e^{\left( -\frac{0.333}{|x_t|}\right)}+0.5 \text{,}\\
x_{t+1}=-3|x_t|^{1.333}+0.5 \text{,}
\end{align}
\end{subequations}

{\setlength{\tabcolsep}{2ex}
\begin{table}[h]
    \centering
    \begin{tabular}{cccc}
        \hline
        
        \fontsize{8}{9}\selectfont\textit{Noise Level} & \fontsize{8}{9}\selectfont\textit{$\sigma = 0.0$} & \fontsize{8}{9}\selectfont\textit{$\sigma = 0.01$} & \fontsize{8}{9}\selectfont\textit{$\sigma = 0.05$}\\
        
        \hline
        
        \fontsize{8}{9}\selectfont\textit{True Model} &  & \fontsize{8}{9}\selectfont\textit{Eq. (5)} & \\ 

         \hline

         \fontsize{8}{9}\selectfont\textit{Expressions} & \fontsize{8}{9}\selectfont\textit{Eq. (8a)} & \fontsize{8}{9}\selectfont\textit{Eq. (8b)} & \fontsize{8}{9}\selectfont\textit{Eq. (8c)}\\ 

         \fontsize{8}{9}\selectfont\textit{Validation MAE} & \fontsize{8}{9}\selectfont\textit{0.00177} & \fontsize{8}{9}\selectfont\textit{0.01236} & \fontsize{8}{9}\selectfont\textit{0.05465}\\ 

         \fontsize{8}{9}\selectfont\textit{RRMSE} & \fontsize{8}{9}\selectfont\textit{0.27\%} & \fontsize{8}{9}\selectfont\textit{2.66\%} & \fontsize{8}{9}\selectfont\textit{10.84\%}\\ 

         \fontsize{8}{9}\selectfont\textit{True RRMSE} & \fontsize{8}{9}\selectfont\textit{0.0\%} & \fontsize{8}{9}\selectfont\textit{2.33\%} & \fontsize{8}{9}\selectfont\textit{12.02\%}\\ 

         \fontsize{8}{9}\selectfont\textit{Convergence} & \fontsize{8}{9}\selectfont\textit{3471} & \fontsize{8}{9}\selectfont\textit{4648} & \fontsize{8}{9}\selectfont\textit{3770}\\ 

         \hline
         
    \end{tabular}
    \caption{\fontsize{9}{10}\selectfont Logistic Map experiment results. Validation MAE excludes regularization and True RRMSE describes the fit of the true model. Convergence is the epoch at which the best Validation MAE is achieved, although losses often saturate well before.}
    \label{tab:my_label}
\end{table}}
\begin{figure}[h]
\centering
\includegraphics[width=\linewidth]{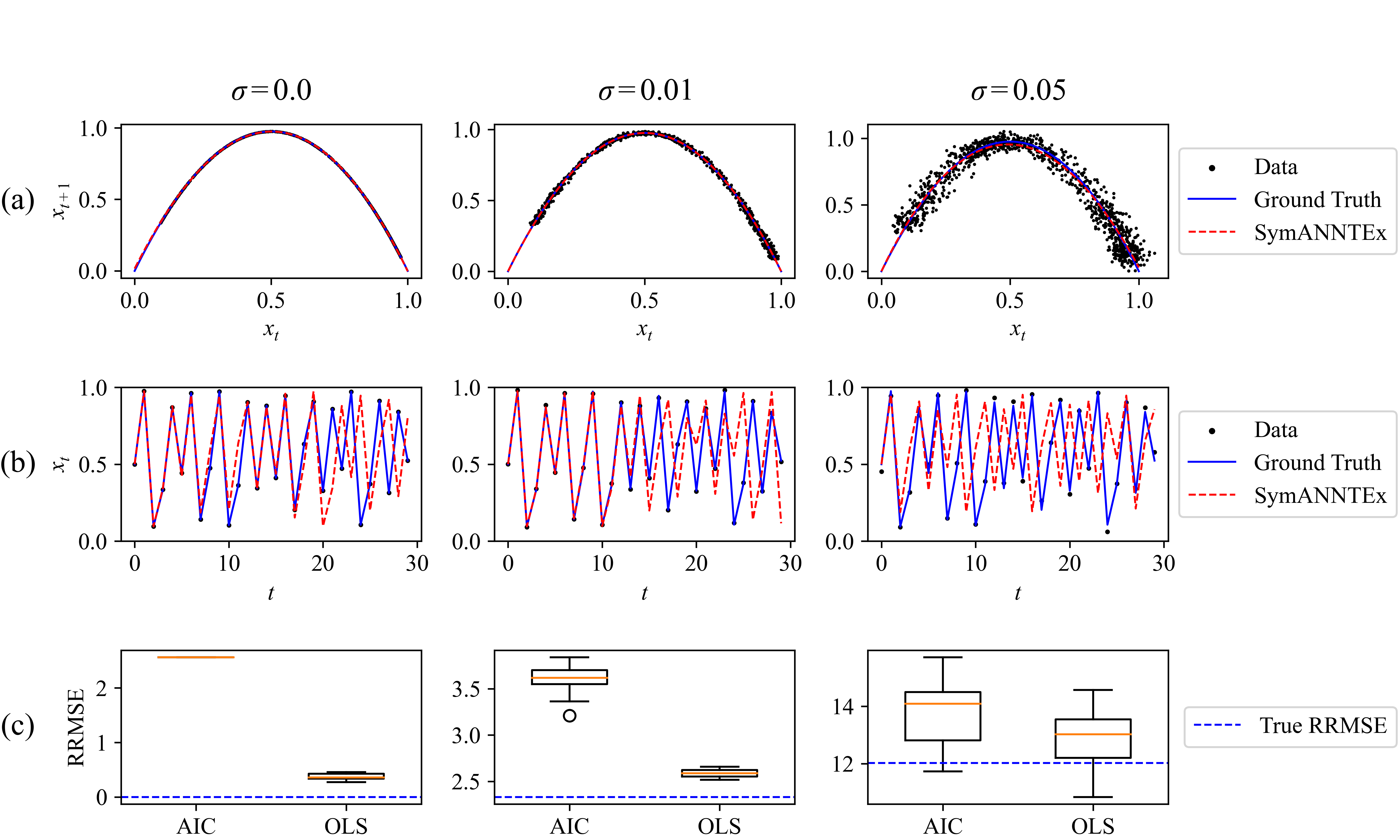}
\caption{\fontsize{9}{10}\selectfont Logistic Map experiment visualizations of (a) state space data and best identified models, (b) trajectories of true and best identified models for 30 iterations, and (c) all RRMSEs at AIC and OLS stages. Note differing vertical axes scales in (c).}
\end{figure}

at $\sigma = 0.0$, $0.01$, and $0.05$, respectively. Further details are given in Table III. Since many other identified models are highly overdetermined, we do not use the OLS refinement stage for this experiment due to computational constraints. Additionally, the OLS stage is unlikely to positively affect Eqs. (9a), (9b), and (9c) since they are incorrectly parameterized and have very high RRMSEs (see Table III).

Moreover, Fig. 2a shows that the identified expressions do not extrapolate well beyond the training domain and Fig. 2b shows that their iterative trajectories rapidly diverge from those of the true model. We consider SymANNTEx on a larger domain of the Gaussian map in Appendix 1, and empirically find that these poor performances are due to the limited training data domain. Thus, data from multiple iterative trajectories or from structured sampling routines may be necessary for accurate system identification.

\subsection*{\fontsize{10}{11}\selectfont
\normalfont \textit{C. Tinkerbell Map}}
Our modified SymANNTEx model identifies
\begin{subequations} 
\begin{align} 
x_{t+1}=0.06x_t y_t - |y_t|^{1.75}+0.05x_t - 0.63y_t \text{,}\\
y_{t+1}=2x_t y_t + 1.93x_t + 0.45y_t - 0.1|y_t|^{1.75}\text{,}\\[10pt]
x_{t+1}=0.39x_t y_t - |y_t|^{1.56}+0.39x_t - 0.5y_t\text{,}\\
y_{t+1}=1.944x_t y_t + 1.944x_t + 0.5y_t\text{,}\\[10pt]
x_{t+1}=-0.08x_t y_t - |y_t|^{1.6}+0.28x_t - 0.43y_t\text{,}\\
y_{t+1}=1.948x_t y_t + 1.889x_t + 0.48y_t\text{,}
\end{align}
\end{subequations}

% TODO

{\setlength{\tabcolsep}{2ex}
\begin{table}[h]
    \centering
    \begin{tabular}{cccc}
        \hline
        
        \fontsize{8}{9}\selectfont\textit{Noise Level} & \fontsize{8}{9}\selectfont\textit{$\sigma = 0.0$} & \fontsize{8}{9}\selectfont\textit{$\sigma = 0.01$} & \fontsize{8}{9}\selectfont\textit{$\sigma = 0.05$}\\
        
        \hline
        
        \fontsize{8}{9}\selectfont\textit{True Model} &  & \fontsize{8}{9}\selectfont\textit{Eq. (6)} & \\ 

         \hline

         \fontsize{8}{9}\selectfont\textit{Expressions} & \fontsize{8}{9}\selectfont\textit{Eq. (9a)} & \fontsize{8}{9}\selectfont\textit{Eq. (9b)} & \fontsize{8}{9}\selectfont\textit{Eq. (9c)}\\ 

         \fontsize{8}{9}\selectfont\textit{Validation MAE} & \fontsize{8}{9}\selectfont\textit{0.00796} & \fontsize{8}{9}\selectfont\textit{0.01201} & \fontsize{8}{9}\selectfont\textit{0.03013}\\ 

         \fontsize{8}{9}\selectfont\textit{RRMSE} & \fontsize{8}{9}\selectfont\textit{113.01\%} & \fontsize{8}{9}\selectfont\textit{81.73\%} & \fontsize{8}{9}\selectfont\textit{82.20\%}\\ 

         \fontsize{8}{9}\selectfont\textit{True RRMSE} & \fontsize{8}{9}\selectfont\textit{0.0\%} & \fontsize{8}{9}\selectfont\textit{2.31\%} & \fontsize{8}{9}\selectfont\textit{11.48\%}\\ 

         \fontsize{8}{9}\selectfont\textit{Convergence} & \fontsize{8}{9}\selectfont\textit{4745} & \fontsize{8}{9}\selectfont\textit{3849} & \fontsize{8}{9}\selectfont\textit{3794}\\ 

         \hline
         
    \end{tabular}
    \caption{\fontsize{9}{10}\selectfont Gaussian Map experiment results. Descriptions of these statistics are in Section III and the caption of Table II.
}
    \label{tab:my_label}
\end{table}}
\begin{figure}[h]
\centering
\includegraphics[width=\linewidth]{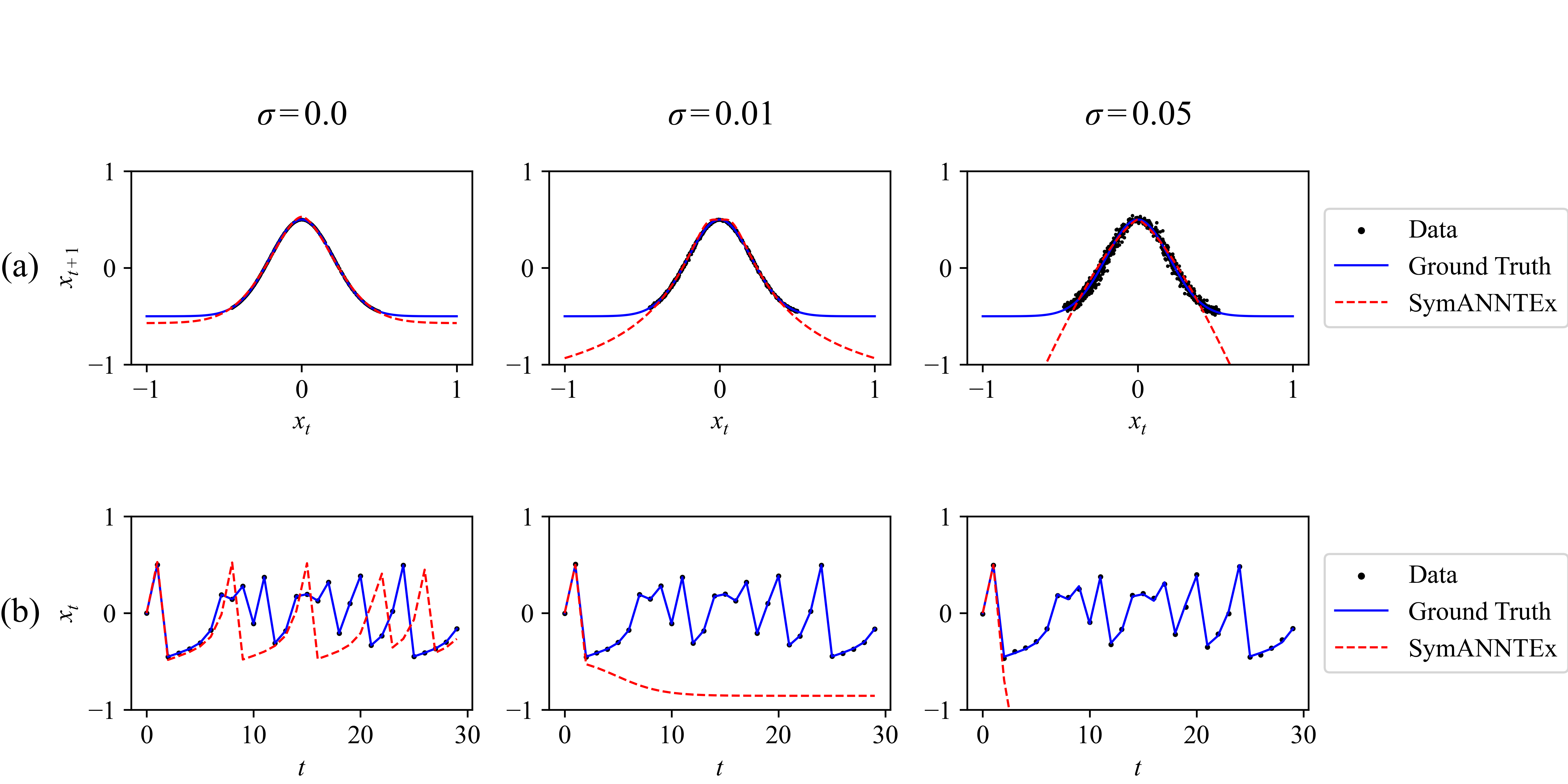}
\caption{\fontsize{9}{10}\selectfont Gaussian Map experiment visualizations of (a) state space data and best identified models and (b) trajectories of true and best identified models for 30 iterations.}
\end{figure}

at $\sigma = 0.0$, $0.01$, and $0.05$, respectively. Further details are given in Table IV. Eqs. (10a), (10c), and (10e) do not appear accurate to the true model, and Fig. 3a shows that these and the true expressions for $x_{t+1}$ have different state-space portraits. However, the negative signomial and fraction of $y_t$ are still captured. We suspect that the absolute-valued nature of the signomials posed a challenge in identifying the signomial of $x_t$ despite our inclusion of the \textit{sign} operator (see Table I). For $y_{t+1}$, Eqs. (10b), (10d), and (10f) are correctly parameterized (except for the small signomial in Eq. (10b)) and reasonably approximate parameters $c$ and $d$. These and the true state-space portraits in Fig. 3b nearly coincide and thus demonstrate accurate identification. Fig. 3c shows that all identified trajectories rapidly diverge from the true model and that the trajectories of Eq. (10a) and (10b) tend towards infinity. However, we find that the trajectories of Eqs. (10c), (10d), (10e), and (10f) remain within the true map’s basin of attraction. Thus, some valuable system behaviors have been identified.

\section*{\centering{\fontsize{11}{12}\selectfont
V. CONCLUSION AND FUTURE WORK}}
We have shown that modified SymANNTEx models (and, more broadly, Deep Learning techniques) are useful tools for iterated map identification. Additionally, we noted the potential utility of preprocessing for noise removal, the importance of sampling techniques for a representative domain, and the possibility of partial identification. Although we consider time-invariant iterated maps in this work due to their interesting properties, we note that SymANNTEx can consider time as an additional term in the input and that our model can also be used for general-purpose symbolic regression. Thus, our work offers a step towards enhanced scientific discovery and interpretation across disciplines and across system types. 

% TODO

{\setlength{\tabcolsep}{0.5ex}
\begin{table}[h]
    \centering
    \begin{tabular}{cccc}
        \hline
        
        \fontsize{8}{9}\selectfont\textit{Noise Level} & \fontsize{8}{9}\selectfont\textit{$\sigma = 0.0$} & \fontsize{8}{9}\selectfont\textit{$\sigma = 0.01$} & \fontsize{8}{9}\selectfont\textit{$\sigma = 0.05$}\\
        
        \hline
        
        \fontsize{8}{9}\selectfont\textit{True Model} &  & \fontsize{8}{9}\selectfont\textit{Eq. (7a), (7b)} & \\ 

         \hline

         \fontsize{8}{9}\selectfont\textit{Expressions} & \fontsize{8}{9}\selectfont\textit{Eq. (10a), (10b)} & \fontsize{8}{9}\selectfont\textit{Eq. (10c), (10d)} & \fontsize{8}{9}\selectfont\textit{Eq. (10e), (10f)}\\ 

         \fontsize{8}{9}\selectfont\textit{Validation MAE} & \fontsize{8}{9}\selectfont\textit{0.04042} & \fontsize{8}{9}\selectfont\textit{0.04064} & \fontsize{8}{9}\selectfont\textit{0.052944}\\ 

         \fontsize{8}{9}\selectfont\textit{RRMSE} & \fontsize{8}{9}\selectfont\textit{60.01\%} & \fontsize{8}{9}\selectfont\textit{60.37\%} & \fontsize{8}{9}\selectfont\textit{205.16\%}\\ 

         \fontsize{8}{9}\selectfont\textit{True RRMSE} & \fontsize{8}{9}\selectfont\textit{0.0\%} & \fontsize{8}{9}\selectfont\textit{1.73\%} & \fontsize{8}{9}\selectfont\textit{8.35\%}\\ 

         \fontsize{8}{9}\selectfont\textit{Convergence} & \fontsize{8}{9}\selectfont\textit{3013} & \fontsize{8}{9}\selectfont\textit{806} & \fontsize{8}{9}\selectfont\textit{3873}\\ 

         \hline
         
    \end{tabular}
    \caption{\fontsize{9}{10}\selectfont Tinkerbell Map experiment results. Descriptions of these statistics are in Section III and the caption of Table II.
}
    \label{tab:my_label}
\end{table}}
\begin{figure}[h]
\centering
\includegraphics[width=\linewidth]{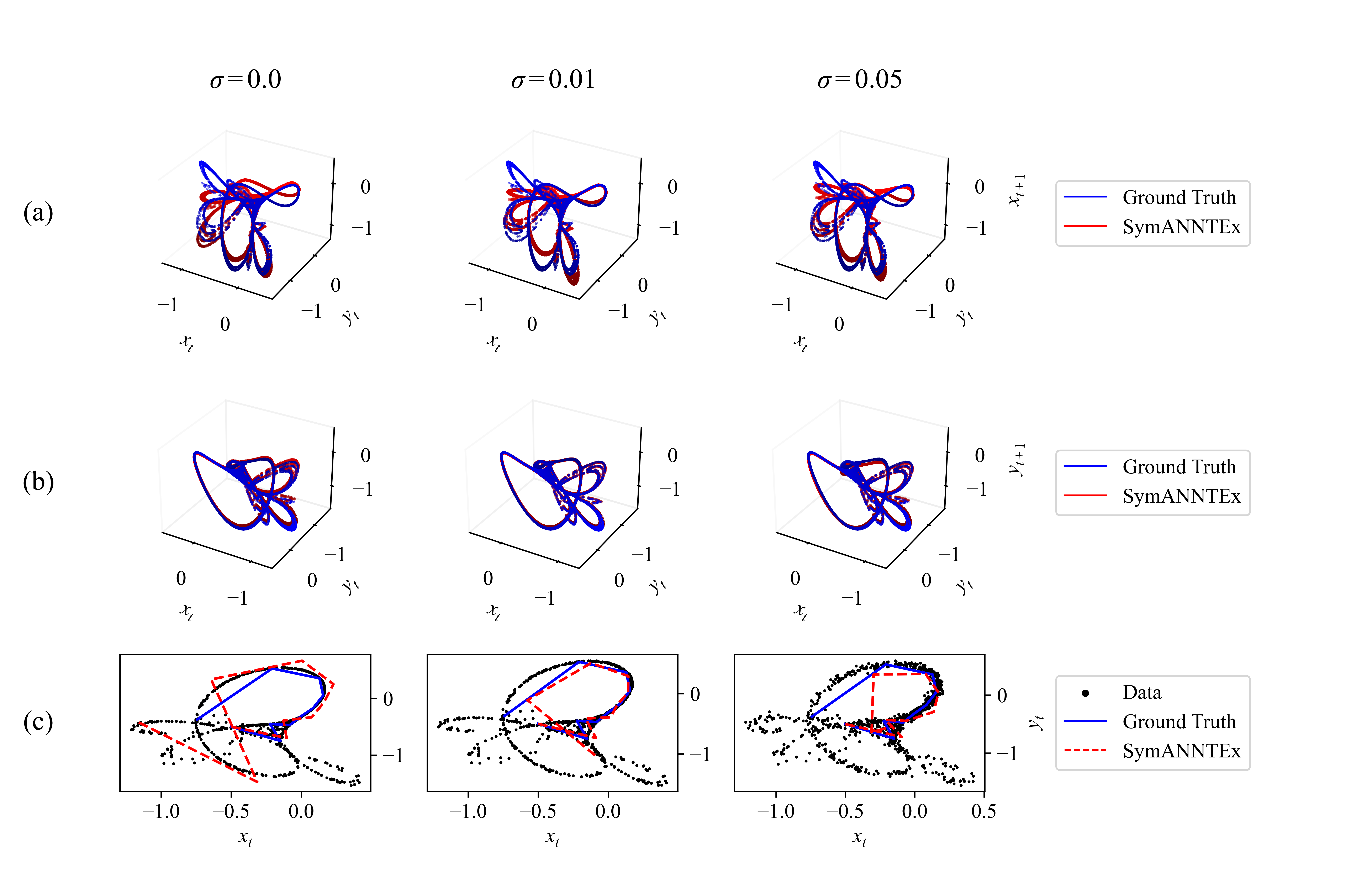}
\caption{\fontsize{9}{10}\selectfont Tinkerbell Map experiment visualizations of (a) three- dimensional state spaces for true and identified $x_{t+1}$ over $x_t$ and $y_t$, (b) similar state spaces for $y_{t+1}$, and (c) true and identified trajectories for 10 iterations. Training data is omitted from the state space plots for clarity.}
\end{figure}

We observe that many signomial weights approximate the correct exponent but are difficult to optimize to the true value. Additionally, we noted in Section IV-C that the absolute-valued nature of the signomial sublayers may be problematic despite the inclusion of a \textit{sign} operator. We leave to future work the further study and principled incorporation of a more robust signomial sublayer. We also encourage further work on enhancements to SymANNTEx, particularly those that facilitate the reproduction of the true model’s statistical properties (\textit{e.g.} first and second order gradients) even when expressions are not entirely accurate.

\section*{\centering{\fontsize{11}{12}\selectfont
ACKNOWLEDGEMENTS}}
The first author would like to thank Dr. Lina Kim, Vamshi Madala, and all others involved in the University of California, Santa Barbara Research Mentorship Program.

\section*{\centering{\fontsize{11}{12}\selectfont
REFERENCES}}
\begin{supertabular}{p{0.04\linewidth} p{0.88\linewidth}}
      [1]  & B. V. Chirikov and V. V. Vecheslavov, “Chaotic dynamics of comet Halley,” \textit{Astron. Astrophys.}, vol. 221, no. 1, pp. 146–154, Aug. 1989.\\
      
      [2] & T. Y Petrosky, “Chaos and cometary clouds in the solar system,” \textit{Phys. Letters A}, vol. 117, no. 7, pp. 328–332, Sep. 1986.\\

      [3]  & F. M. Izraelev, “Nearly linear mappings and their applications,” \textit{Phys. D: Nonlin. Phen.}, vol. 1, no. 3, pp. 243–266, Sep. 1986.\\

      [4]  &  D. W. Crevier and M. Meister, “Synchronous Period-Doubling in Flicker Vision of Salamander and Man,” \textit{J. Neurophysio.}, vol. 79, no. 4, pp. 1869–1878, Apr. 1998.\\

      [5]  & Z. Qu, “Chaos in the genesis and maintenance of cardiac arrhythmias,” \textit{Prog. Biophys. \& Molec. Bio.}, vol. 105, no. 3, pp. 247–257, May 2011.\\

      [6]  &  P. G. Reinhall, T. K. Caughey, and D. W. Storti, “Order and Chaos in a Discrete Duffing Oscillator: Implications on Numerical Integration,” \textit{J. Applied Mech.}, vol. 56, no. 1, pp. 162–167, Mar. 1983.\\

      [7]  & N. Boddupalli, T. Matchen, and J. Moehlis, “Symbolic Regression via Neural Networks,” \textit{Chaos: Interdisc. J. Nonlin. Sci.}, vol. 33, no. 8, Aug. 2023.\\

      [8]  & S. L. Brunton, J. L. Proctor, and J. N. Kutz, “Discovering governing equations from data via sparse identification of nonlinear dynamical systems,” \textit{PNAS}, vol. 113, no. 15, pp. 3932–3937, Mar. 2016.\\

      [9]  & C. Fronk and L. Petzold, “Interpretable polynomial neural ordinary differential equations,” \textit{Chaos: Interdisc. J. Nonlin. Sci.}, vol. 33, no. 4, Apr. 2023.\\

      [10]  & G. G. Chrysos, S. Moschoglou, G. Bouritsas, Y. Panagakis, J. Deng, and S. Zafeiriou, “$\Pi$-nets: Deep Polynomial Neural Networks,” \textit{Proc. CVPR}, Jun. 2020.\\

      [11]  & R. T. Q. Chen, Y. Rubanova, J. Bettencourt, and D. K. Duvenaud, “Neural Ordinary Differential Equations,” \textit{Proc. NeurIPS}, vol. 31, Dec. 2018.\\

      [12]  & G. Martius and C. H. Lampert, “Extrapolation and learning equations,” \textit{arXiv:1610.02995}, Oct. 2016.\\

      [13]  & Z. Xu, H. Zhang, Y. Wang, X. Chang, and Y. Liang, “$L_{1/2}$ regularization,” \textit{Sci. China Inf. Sci.}, vol. 53, no. 6, pp. 1159–1169, Jun. 2010.\\

      [14]  & D. P. Kingma, J. Ba, “Adam: A Method for Stochastic Optimization,” \textit{Proc. ICLR}, Dec. 2014.\\

      [15]  & L. N. Smith, "Cyclical Learning Rates for Training Neural Networks," \textit{Proc. IEEE WACV}, pp. 464-472, Mar. 2017.\\

      [16]  & A. Meurer \textit{et al.}, “SymPy: symbolic computing in Python,” \textit{PeerJ Computer Science}, vol. 3, Jan. 2017.\\

      [17]  & H. Akaike, “Information Theory and an Extension of the Maximum Likelihood Principle”, in \textit{Selected Papers of Hirotugu Akaike}, E. Parzen, K. Tanabe, and G. Kitagawa, Eds. New York, NY: Springer New York, 1998, pp. 199–213.\\

      [18]  & A. Goldsztejn, W. Hayes, and P. Collins, “Tinkerbell Is Chaotic”, \textit{SIAM J. Applied Dyn. Sys.}, vol. 10, no. 4, pp. 1480–1501, Dec. 2011.\\
\end{supertabular}

\section*{\centering{\fontsize{11}{12}\selectfont
APPENDIX}}
\subsection*{\centering{\fontsize{10}{11}\selectfont
1. Gaussian Map on a Larger Domain}}

{\setlength{\tabcolsep}{2ex}
\begin{table}[h]
    \centering
    \begin{tabular}{cccc}
        \hline
        
        \fontsize{8}{9}\selectfont\textit{Noise Level} & \fontsize{8}{9}\selectfont\textit{$\sigma = 0.0$} & \fontsize{8}{9}\selectfont\textit{$\sigma = 0.01$} & \fontsize{8}{9}\selectfont\textit{$\sigma = 0.05$}\\
        
        \hline
        
        \fontsize{8}{9}\selectfont\textit{True Model} &  & \fontsize{8}{9}\selectfont\textit{Eq. (6)} & \\ 

         \hline

         \fontsize{8}{9}\selectfont\textit{Expressions} & \fontsize{8}{9}\selectfont\textit{Eq. (11a)} & \fontsize{8}{9}\selectfont\textit{Eq. (11b)} & \fontsize{8}{9}\selectfont\textit{Eq. (11c)}\\ 

         \fontsize{8}{9}\selectfont\textit{Validation MAE} & \fontsize{8}{9}\selectfont\textit{0.00681} & \fontsize{8}{9}\selectfont\textit{0.00899} & \fontsize{8}{9}\selectfont\textit{0.03021}\\ 

         \fontsize{8}{9}\selectfont\textit{RRMSE} & \fontsize{8}{9}\selectfont\textit{2.70\%} & \fontsize{8}{9}\selectfont\textit{3.03\%} & \fontsize{8}{9}\selectfont\textit{12.07\%}\\ 

         \fontsize{8}{9}\selectfont\textit{True RRMSE} & \fontsize{8}{9}\selectfont\textit{0.0\%} & \fontsize{8}{9}\selectfont\textit{2.38\%} & \fontsize{8}{9}\selectfont\textit{11.25\%}\\ 

         \fontsize{8}{9}\selectfont\textit{Convergence} & \fontsize{8}{9}\selectfont\textit{4026} & \fontsize{8}{9}\selectfont\textit{4334} & \fontsize{8}{9}\selectfont\textit{4259}\\ 

         \hline
         
    \end{tabular}
    \caption{\fontsize{9}{10}\selectfont Larger-domain Gaussian Map experiment results. Descriptions of these statistics are in Section III and the caption of Table II.}
    \label{tab:my_label}
\end{table}}
\begin{figure}[h]
\centering
\includegraphics[width=\linewidth]{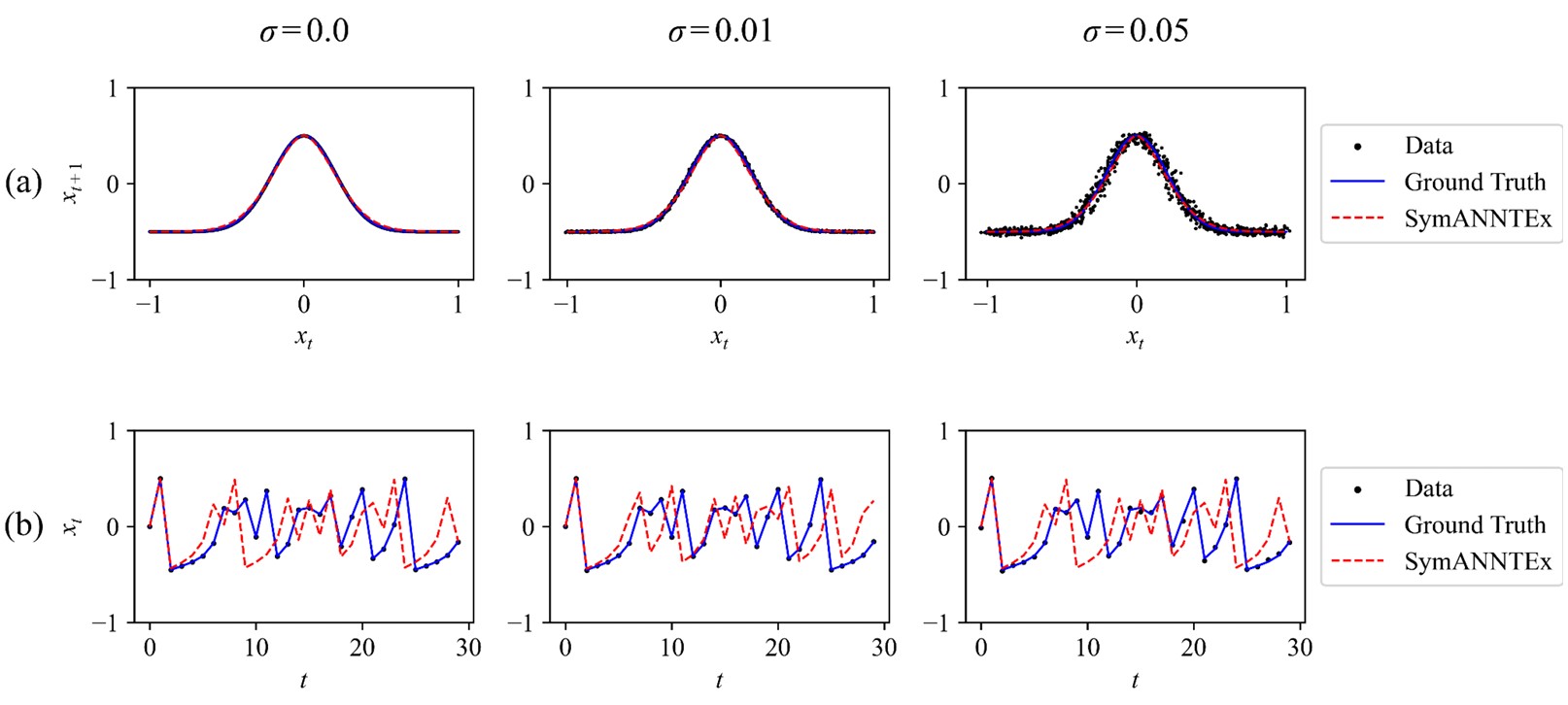}
\caption{\fontsize{9}{10}\selectfont Larger-domain Gaussian Map experiment visualizations of (a) state space data and best identified models and (b) trajectories of true and best identified models.}
\end{figure}

We observe in Section III-C that SymANNTEx fails to identify properly extrapolating models when trained on a single limited-domain iterative trajectory where the tails are not represented in the data. To consider the effects of the limited domain, we perform an additional experiment where the training data is linearly spaced on the larger domain of $[-1.0, 1.0]$ before perturbation. The experiment is otherwise conducted identically to the original in Section III-C. SymANNTEx accordingly identifies
\begin{subequations} 
\begin{align} 
x_{t+1}=e^{\left( -9.6377|x_t|^{1.8392}\right)}-0.5027 \text{,}\\
x_{t+1}=e^{\left( -10.1898|x_t|^{1.8798}\right)}-0.5019 \text{,}\\
x_{t+1}=e^{\left( -8.4013|x_t|^{1.736}\right)}-0.5043 \text{,}
\end{align}
\end{subequations}

at $\sigma = 0.0$, $0.01$, and $0.05$, respectively. Further details are given in Table V. We conduct the OLS refinement on only these best identified models since many instances yield AIC-simplified models that are highly overparameterized. The OLS-refined models — Eqs. (11a), (11b), and (11c) — are all of the correct form, identify $\beta$ very closely, and reasonably approximate $\alpha$ and the signomial order. Additionally, their RRMSEs are dramatically lower than those in the single-trajectory experiment (see Table IV). As compared to the $\sigma = 0.0$ and $\sigma = 0.01$ cases for the Gaussian Map and Logistic Map, we observe similarly reduced signomial orders in the $\sigma = 0.05$ cases Eq. (11c) and Eq. (8c). However, Table V shows that Eq. (11c) does not achieve a RRMSE lower than that of the true model. Thus, SymANNTEx appears to be more robust to noise on the larger-domain Gaussian Map (\textit{i.e.} identified expressions do not fit the noisy data better than the true model does). This phenomenon is not due to relative noise scales on a larger domain as our perturbations are scaled by the RMS of the state variable. Therefore, due to SymANNTEx’s enhanced performance on larger domains, multiple iterative trajectories or structured sampling routines may be required for accurate system identification.

\end{document}